\documentclass[10pt,twocolumn,letterpaper]{article}
\PassOptionsToPackage{table}{xcolor}

\usepackage{cvpr}              
\usepackage{cuted, capt-of}
\usepackage[table]{xcolor}
\definecolor{cvprblue}{rgb}{0.21,0.49,0.74}
\usepackage[pagebackref,breaklinks,colorlinks,allcolors=cvprblue]{hyperref}
\usepackage{graphicx}
\usepackage{booktabs}
\usepackage{multirow}

\def\paperID{2873} 
\def\confName{CVPR}
\def\confYear{2026}

\title{
    FSMC-Pose: Frequency and Spatial Fusion with Multiscale Self-calibration for Cattle Mounting Pose Estimation
}

\author{
Fangjing Li$^{1}$ \quad
Zhihai Wang$^{1}$ \quad
Xinxin Ding$^{1,2}$ \quad
Haiyang Liu$^{1}$\footnotemark[1] \quad\\
Ronghua Gao$^{2}$ \quad
Rong Wang$^{2}$ \quad
Yao Zhu$^{3}$\footnotemark[1] \quad
Ming Jin$^{4}$\\
$^{1}$Beijing Jiaotong University \quad
$^{2}$NERCITA \quad
$^{3}$Tsinghua University \quad
$^{4}$Griffith University\\
}

\begin{document}

\maketitle

\footnotetext[1]{Correspondence to: Haiyang Liu \textless haiyangliu@bjtu.edu.cn\textgreater\ and Yao Zhu \textless ee\_zhuy@zju.edu.cn\textgreater.}

\begin{abstract}
Mounting posture is an important visual indicator of estrus in dairy cattle. However, achieving reliable mounting pose estimation in real-world environments remains challenging due to cluttered backgrounds and frequent inter-animal occlusion. We present \textbf{FSMC-Pose}, a top-down framework that integrates a lightweight frequency–spatial fusion backbone, CattleMountNet, and a multiscale self-calibration head, SC2Head. Specifically, we design two algorithmic components for CattleMountNet: the Spatial Frequency Enhancement Block (SFEBlock) and the Receptive Aggregation Block (RABlock). SFEBlock separates cattle from cluttered backgrounds, while RABlock captures multiscale contextual information. The Spatial-Channel Self-Calibration Head (SC2Head) attends to spatial and channel dependencies and introduces a self-calibration branch to mitigate structural misalignment under inter-animal overlap. We construct a mounting dataset, MOUNT-Cattle, covering 1{,}176 mounting instances, which follows the COCO format and supports drop-in training across pose estimation models. Using a comprehensive dataset that combines MOUNT-Cattle with the public NWAFU-Cattle dataset, FSMC-Pose achieves higher accuracy than strong baselines, with markedly lower computational and parameter costs, while maintaining real-time inference on commodity GPUs. Extensive experiments and qualitative analyses show that FSMC-Pose effectively captures and estimates cattle mounting pose in complex and cluttered environments. Dataset and code are available at \href{https://github.com/elianafang/FSMC-Pose}{Github}.

\end{abstract}

\begin{figure}[h]
\centering
\includegraphics[width=\linewidth]{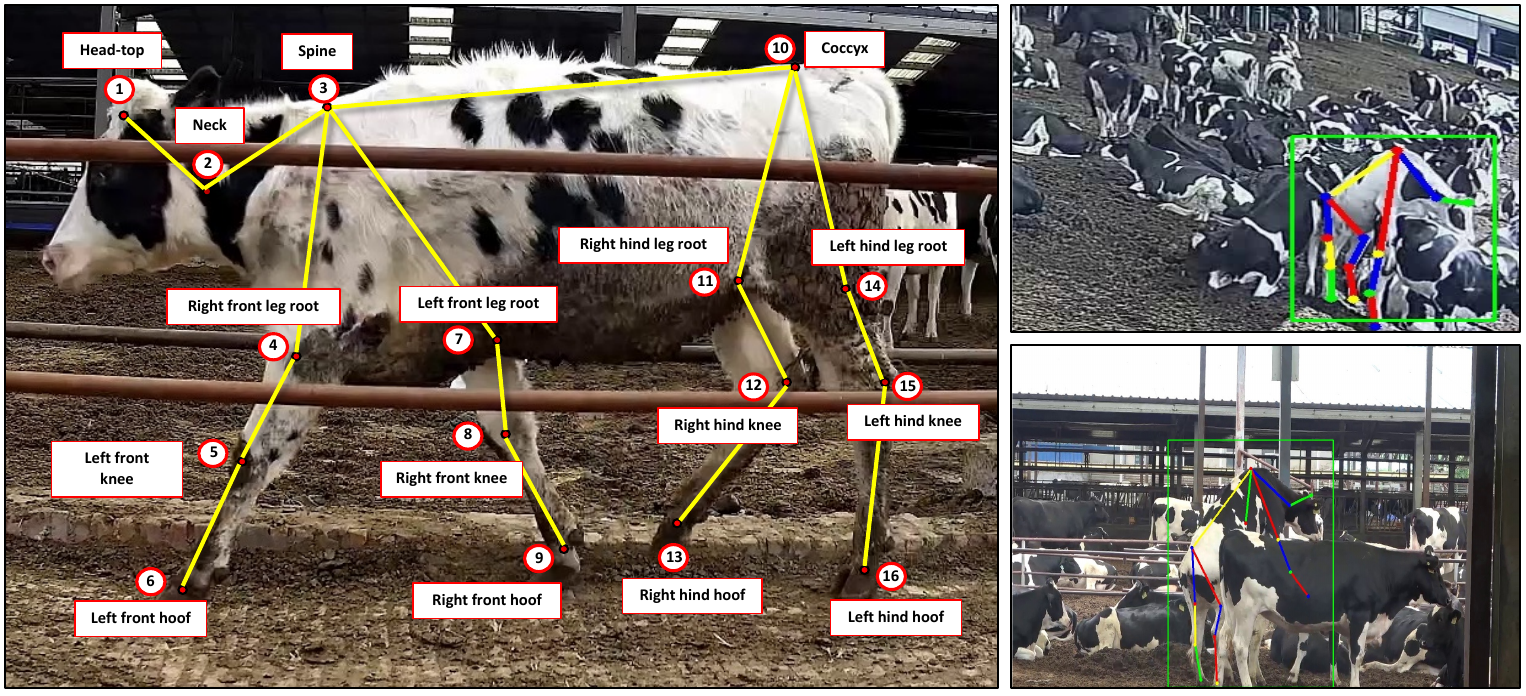}
\caption{\textbf{Left}: keypoint annotation scheme for cattle. \textbf{Right}: real-world dense environments for cattle mounting pose estimation by FSMC-Pose, based on self-collected MOUNT-Cattle dataset.}
\label{fig:Keypoint}
\end{figure}

\section{introduction}
Accurate estrus identification is pivotal to herd profitability and sustainability, influencing conception timing, days open, calving interval, labor cost, hormone use, and animal welfare~\cite{van1996sexual}. Among behavioral indicators, mounting is the most intuitive and visually distinctive pose, characterized by forelimb lifting and hindlimb support, and it provides a critical behavioral cue for determining whether a cattle has entered estrus. If mounting pose can be measured automatically on a large scale, closed-loop decision-making in breeding, resource allocation, and health monitoring can be achieved. A reliable mounting pose estimation model would convert ubiquitous low-cost video into actionable signals, lowering dependence on skilled labor, improving reproductive efficiency, and reducing waste across diverse farm conditions. However, there is a lack of public cattle mounting datasets, resulting in the lack of research foundation for this agricultural production problem. Consequently, cattle mounting pose estimation research remains a blank.

Pose estimation~\cite{wang2022contextual,an2024sharpose,hu2024continuous,wang2024gtpt,wang2024spatial} provides structured visual perception by extracting anatomical keypoints and spatial topology for reliable behavior recognition~\cite{mehmood2025extended,xie2024dynamic,zhou2024blockgcn,xu2025language}. Current animal pose estimation methods mainly follow bottom-up~\cite{artacho2023bapose,yang2024x,qu2023characteristic} or top-down~\cite{li2022cliff,khirodkar2021multi,yin2024srpose} paradigms. Bottom-up methods such as DeepLabCut~\cite{mathis2018deeplabcut} effectively transfer human pose architectures to animals with limited annotations but fail under occlusion and inter-animal confusion. GANPose~\cite{wang2023ganpose} introduces structural priors for occluded inference but is computationally expensive, while CMBN~\cite{fan2023bottom} reduces parameters by HRNet~\cite{wang2020deep} optimization yet still struggles in dense herd scenes. More importantly, agricultural production requires real-time monitoring and feedback, but the high computational cost of bottom-up approaches further limits their adoption in real-time production scenarios. Top-down methods like GRMPose~\cite{chen2024grmpose} and T-LEAP~\cite{russello2022t} improve accuracy through lightweight backbones or temporal modeling but suffer from keypoint obfuscation and high inference complexity.

Existing approaches largely transfer human pose estimation models to animals, but the complexity of agricultural production scenes makes these methods unsuitable for real-world deployment (Figure~\ref{fig:Keypoint}). Specifically, estrous cattle tend to aggregate, making mounting scenes denser than typical farm settings. Therefore, cluttered background interference and occlusion by other cattle blur or partially remove the mounting outline; in crowded views, it is difficult to fully distinguish the individuals involved in mounting, with intertwined limbs and joints causing identity confusion; moreover, overlapping coat patterns further increase the difficulty of keypoint recognition during pose estimation. So the question of ``\textit{\textbf{how to improve mounting pose estimation in dense, cluttered real herd scenes while maintaining lightweight computation?}}'' remains a challenge.

To address these issues, we propose \textbf{FSMC-Pose}, a frequency and spatial fusion framework with multiscale self-calibration for cattle mounting pose estimation in real-world group-housed environments. Frequency and spatial fusion uses wavelet decomposition and fixed-Gaussian smoothing to suppress clutter, enhance separability between cattle and background, and preserve fine structural detail at low contrast. Multiscale self-calibration utilizes receptive field aggregation and spatial–channel co-calibration to aggregate context across scales, correct structural shifts under inter-animal overlap, and stabilize keypoint localization for small joints and large torso regions. Extensive experiments show that FSMC-Pose accurately captures mounting postures in complex scenes and provides an effective technological foundation for intelligent estrus detection systems. Our main contributions are summarized as follows:

\begin{itemize}
\item We propose FSMC-Pose, a lightweight top-down framework integrating a novel backbone CattleMountNet and SC2Head for robust mounting pose estimation in real-world, group-housed dairy cattle environments.
\item We construct MOUNT-Cattle dataset and combine it with NWAFU-Cattle~\cite{fan2023bottom} to form a comprehensive benchmark for complex mounting environments. The annotations follow the COCO~\cite{lin2014microsoft} format and support drop-in training across pose estimation models, covering 1{,}176 mounting instances (Figure~\ref{fig:Keypoint}).

\item Extensive experiments show FSMC-Pose surpasses strong baselines on cattle mounting pose estimation tasks, while maintaining real-time inference on commodity GPUs. Specifically, FSMCPose improves AP, AP\textsubscript{75}, AR, and AR\textsubscript{75} by 1.4\%, 3.0\%, 0.9\%, and 0.4\%, reaching 89\%, 92.5\%, 89.9\%, and 97.7\%, respectively. Compared with RTMPose~\cite{jiang2023rtmpose}, its computational cost is only 4.4109 GFLOPS, and its parameter count is reduced by 80.01\% to just 2.698M.
\end{itemize}

\section{Related Work}

Pose estimation~\cite{wang2022contextual,an2024sharpose,hu2024continuous,wang2024spatial,yang2023effective} as a structured visual perception method allows us to extract keypoints and spatial topology, providing reliable intermediate representations for behavior recognition~\cite{mehmood2025extended,xie2024dynamic,zhou2024blockgcn,xu2025language}. With the rapid development of computer vision, animal pose estimation has also made significant progress. Existing methods can generally be divided into bottom-up~\cite{artacho2023bapose,yang2024x,qu2023characteristic} and top-down~\cite{li2022cliff,khirodkar2021multi,yin2024srpose} paradigms. 

\subsection{Bottom-up Methods} Existing bottom-up methods extend human pose estimators to animal data. DeepLabCut~\cite{mathis2018deeplabcut} fine-tunes human architectures on a small number of annotated samples, reducing labeling costs but showing limited ability to distinguish individuals in crowded scenes and being vulnerable to occlusion and feature confusion. GANPose~\cite{wang2023ganpose} introduces a generative adversarial network with structural priors to infer occluded poses without temporal information, yet requires substantial computation and large, high-quality annotations, hindering deployment in farms. CMBN~\cite{fan2023bottom} compresses the HRNet backbone~\cite{wang2020deep} with depthwise separable convolutions, but still mis-associates keypoints across individuals in dense production scenes, and the overall computational cost of bottom-up pipelines constrains real-time monitoring.

\subsection{Top-down Methods} Top-down methods generally use fewer parameters and achieve higher keypoint accuracy. GRMPose~\cite{chen2024grmpose} couples the lightweight CSPNext backbone~\cite{lyu2022rtmdet} with a graph convolutional coordination classifier to balance speed and accuracy, yet similar coat patterns and ambiguous body contours in herds still cause structural confusion and missing parts. Video-based work~\cite{russello2024video} extends LEAP~\cite{pereira2019fast} to the temporal model T-LEAP~\cite{russello2022t}, enlarging the receptive field via sequential frames, but it cannot infer pose from a single frame and incurs high inference complexity. Overall, existing approaches target relatively simple, low-overlap scenes and provide little dedicated support for mounting pose estimation.

\begin{figure*}[t]
\centering
\includegraphics[width=0.92\textwidth]{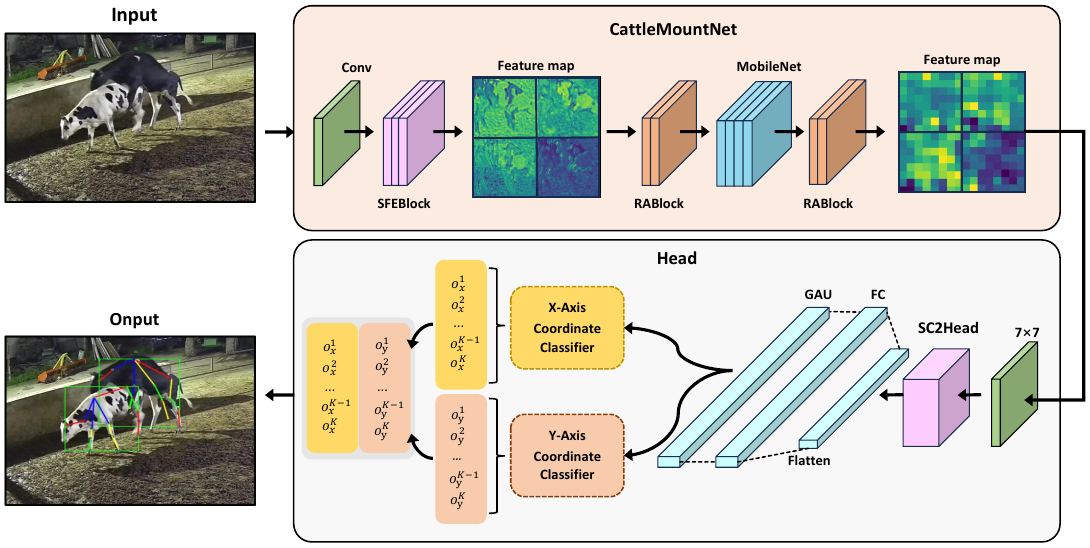}
\caption{Architecture of FSMC-Pose framework, including the proposed lightweight backbone CattleMountNet (SFEBlock and RABlock) (Figure~\ref{fig:SFE_RA_blocks}) and self-calibration SC2Head (Figure~\ref{fig:SC2Head}). Following the top-down design of RTMPose~\cite{jiang2023rtmpose}, and employing MobileNet~\cite{sandler2018mobilenetv2}.}
\label{fig:FSMC-Pose}
\end{figure*}

\section{Dataset}

We constructed a mounting dataset called MOUNT-Cattle, which is a mounting-centric dairy cattle pose dataset collected from real-world farms. Specifically, MOUNT-Cattle was recorded at a large commercial dairy farm in Yanqing District, Beijing, during July to August 2024, using an infrared network camera (Hikvision DS-2CD3T46WDV3-I3) and a Sony FDR-AX60 4K camcorder. The dataset focuses on mounting behavior under dense herd conditions, deliberately covering severe background clutter, similar coat patterns, and mutual occlusion, and preserving the full mounting process from initiation to termination. After manual filtering, MOUNT-Cattle contains 1{,}176 high-quality annotated mounting instances. We then combine our self-collected MOUNT-Cattle dataset with the public NWAFU-Cattle dataset~\cite{fan2023bottom}, which does not include mounting behavior, to construct a comprehensive benchmark dataset.

\paragraph{Dataset Split.}
Each cattle instance in this combined dataset is labeled in COCO format~\cite{lin2014microsoft} with a bounding box and 16 keypoints following Animal-Pose~\cite{cao2019cross} and AP-10K~\cite{yu2021ap}, while omitting eye and mouth keypoints and adding head-top and neck to focus on whole-body pose. Keypoint visibility is categorized as invisible, partially visible, or visible, and the data are split into train/validation/test sets with an $8\!:\!1\!:\!1$ ratio. More details of MOUNT-Cattle and the combined benchmark are provided in appendix \textcolor{cvprblue}{A}.

\section{Methodology}
\subsection{Overview}
FSMC-Pose comprises a lightweight backbone network, CattleMountNet, and an improved pose estimation head, SC2Head. As shown in Figure~\ref{fig:FSMC-Pose}, the input image is normalized and then processed by the backbone network to extract multi-level features. For CattleMountNet, we integrate depthwise separable convolutions, residual connections, and inverted residual structures in a modular fashion to enhance the model’s capability for foreground–background discrimination and scale representation.

To further feature extraction, we design two modules, SFEBlock and RABlock, from a complementary perspective. They fuse frequency and spatial domain information and by modeling multiscale receptive fields, respectively. Following feature extraction, we employ a spatial–channel self-calibration mechanism to focus the attention on critical body regions, and adopt a coordinate regression strategy for keypoint prediction. Furthermore, we introduce the SC2Head to enhance feature representations during keypoint localization based on RTMPose~\cite{jiang2023rtmpose}.

\begin{table}[t]
\centering
\caption{Statistics of keypoint visibility categories across dataset splits. Values are counts with proportions in parentheses.}
\resizebox{\columnwidth}{!}{
\begin{tabular}{ccccc}
\toprule
\textbf{Split} & \textbf{Invisible} & \textbf{Partial} & \textbf{Visible} & \textbf{Total} \\
\midrule
Train & 7,493 (11.51\%) & 4,646 (7.14\%) & 52,965 (81.35\%) & 65,104 \\
Val   & 929 (11.14\%)   & 564 (6.77\%)   & 6,843 (82.09\%)  & 8,336 \\
Test  & 1,066 (12.81\%) & 572 (6.88\%)   & 6,682 (80.31\%)  & 8,320 \\
\midrule
Total & 9,488 (11.60\%) & 5,782 (7.07\%) & 66,490 (81.32\%) & 81,760 \\
\bottomrule
\end{tabular}
}
\label{tab:visibility}
\end{table}

\subsection{Lightweight Backbone: CattleMountNet}
We build CattleMountNet on inverted residual structures~\cite{howard2019searching,mehta2022separable}: a feature map of size $H \times W \times C$ is first expanded by a $1 \times 1$ pointwise convolution, processed by a $3 \times 3$ depthwise convolution, then projected back to low dimension with another $1 \times 1$ pointwise convolution and fused with the input. This bottleneck design preserves key information while keeping computation low. To better handle dense, cluttered group-housed cattle scenes, we introduce two modules on top of this structure: the Spatial-Frequency Enhancement Block (SFEBlock) and the Receptive Aggregation Block (RABlock). SFEBlock enhances separation between cattle and background via frequency–spatial modeling, and RABlock aggregates multiscale context to handle strong keypoint scale variation.

\begin{figure}[t]
\centering
\begin{minipage}{\linewidth}
    \centering
    \includegraphics[width=\linewidth]{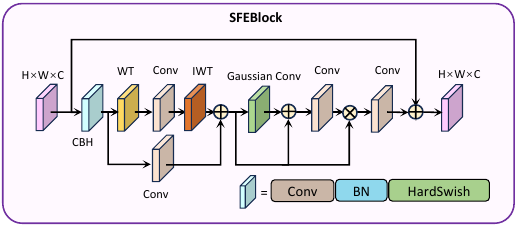}
\end{minipage}\hfill
\begin{minipage}{\linewidth}
    \centering
    \includegraphics[width=\linewidth]{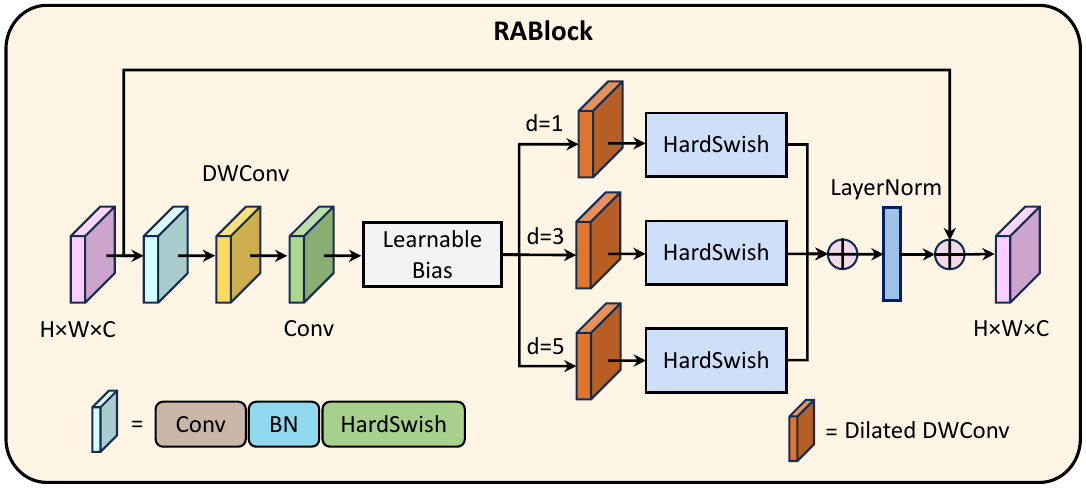}
\end{minipage}
\caption{Architectures of the proposed CattleMountNet components: SFEBlock and RABlock.}
\label{fig:SFE_RA_blocks}
\end{figure}

\paragraph{Spatial-Frequency Enhancement Block (SFEBlock).}
In real barns, mud, shadows, and lighting often make cow textures similar to the background, causing low contrast and blurred keypoints that degrade with depth. SFEBlock is designed to strengthen target–background separation while remaining lightweight, as illustrated in Figure~\ref{fig:SFE_RA_blocks}.

SFEBlock combines Wavelet Transform Convolution (WTConv)~\cite{finder2024wavelet} and Gaussian filtering. Wavelets provide multiscale frequency-domain modeling with enlarged receptive fields, while the Gaussian kernel smooths responses and suppresses background noise. Given input $F_{in} \in \mathbb{R}^{H \times W \times C}$, we first decompose it with WT and convolve each sub-band:
\begin{equation}
F_{WTconv} = \operatorname{IWT}\left(\operatorname{Conv}\left(W, \operatorname{WT}\left(F_{in}\right)\right)\right),
\label{eq:wtconv}
\end{equation}
where $W$ is the depthwise kernel for wavelet sub-bands. WT is downsampled low- and high-frequency components; small kernels on each band capture context while preserving local structure, and IWT reconstructs spatial features.

Pixels near the center receive higher weights, emphasizing salient structure and suppressing noise. We use a fixed $5 \times 5$ kernel, $F_{\text{gauss}} = G_{1.0}^{5 \times 5}\left(F_{WTconv}\right)$, to smooth each channel, then fuse wavelet and Gaussian features and compress them via a $1 \times 1$ convolution; element-wise multiplication and a $3 \times 3$ convolution further refine spatial responses, and a residual connection preserves the input, yielding:
\begin{align}
F_{\text{temp}} &= \operatorname{Conv}_{2D}^{1 \times 1}\left(F_{WTconv} + F_{\text{gauss}}\right),
\label{eq:ftemp}\\
F_{\text{out}} &= \operatorname{Conv}_{2D}^{3 \times 3}\left(F_{WTconv} \otimes F_{\text{temp}}\right) + F_{in}.
\label{eq:fout}
\end{align}
This improves contrast on cattle contours while keeping computation modest.

\paragraph{Receptive Aggregation Block (RABlock).}
Cattle keyparts span small hooves and large torso or spine regions. Single-scale features cannot simultaneously capture such variation in cluttered scenes. RABlock addresses this via parallel depthwise convolutions with different dilation rates plus residual aggregation, as shown in Figure~\ref{fig:SFE_RA_blocks}.

On top of the inverted residual unit, we add learnable channel-wise biases for lightweight distribution adjustment. The main branch contains three parallel $3 \times 3$ depthwise convolutions with dilation rates $1$, $3$, and $5$, capturing local, mid-range, and long-range context. For input $\mathrm{F}_{l-1} \in \mathbb{R}^{H \times W \times C}$ at layer $L$, RABlock is defined as:
\begin{align}
\mathbf{H}_{l-1}^{n} &= \operatorname{HardSwish}\left(\operatorname{Conv}_{3 \times 3}^{\mathrm{dil}=2n-1}\left(\mathrm{F}_{l-1}\right)\right), \\
\mathbf{H}_{l-1} &= \operatorname{LayerNorm}\left(\mathbf{H}_{l-1}^{1} + \mathbf{H}_{l-1}^{2} + \mathbf{H}_{l-1}^{3}\right).
\label{eq:H_l-1}
\end{align}
Depthwise convolutions keep parameters low, while HardSwish~\cite{howard2019searching} provides efficient nonlinearity for mobile settings. Summing and normalizing the three paths yields a multiscale feature map that better responds to both small joints and large body structures. Residual connections in each path help preserve original structure and stabilize training under strong scale and background variation.

\subsection{Multiscale Self-calibration Head: SC2Head}
\begin{figure}[t]
\centering
\includegraphics[width=\linewidth]{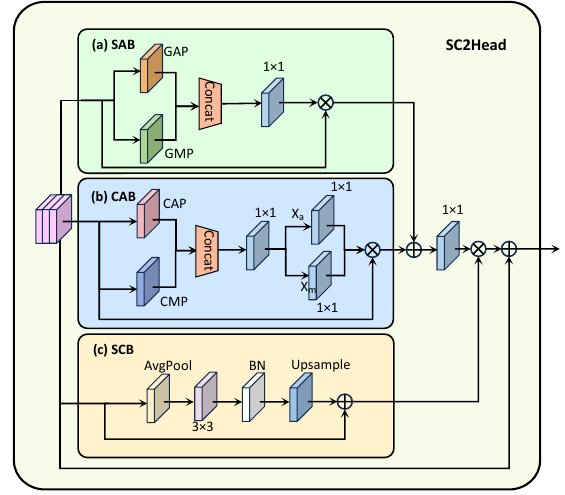}
\caption{Architecture of the proposed SC2Head module for spatial–channel co-calibrated keypoint prediction. The improved visualization is shown in Figure~\ref{fig:Heatmaps}.}
\label{fig:SC2Head}
\end{figure}

In group-housed mounting scenes, similar coat patterns and strong inter-cow overlap make keypoints of the same body parts spatially close and semantically ambiguous, causing structural confusion and mis-association between individuals. Our backbone with SFEBlock and RABlock improves cow–background separation and multiscale representation, but these effects mainly act in early feature extraction, and the prediction head still struggles to maintain structural consistency. To address this, we introduce SC2Head on top of RTMPose~\cite{jiang2023rtmpose}, as shown in Figure~\ref{fig:SC2Head}, which couples spatial and channel attention with a self-calibration branch to correct structural shifts and enhance keypoint localization.

The SC2Head consists of three branches: the Spatial Attention Branch (SAB), the Channel Attention Branch (CAB), and the Self-Calibration Branch (SCB). Given an input feature $\mathrm{X} \in \mathbb{R}^{H \times W \times C}$, SC2Head is defined as:
\begin{align}
\mathbf{C}_{o} &= f_{1 \times 1}([\mathrm{SAB}(\mathbf{X}), \mathrm{CAB}(\mathbf{X})] \odot \mathrm{SCB}(\mathbf{X}))+\mathbf{X} \\
&= f_{1 \times 1}([\mathrm{SA}, \mathrm{CA}]) \odot \mathrm{SC}+\mathbf{X},
\label{eq:SC2Head}
\end{align}
where $\mathbf{C}_{o} \in \mathbb{R}^{H \times W \times C}$ denotes the SC2Head output, $f_{1 \times 1}$ represents a $1 \times 1$ convolution, $\odot$ denotes the broadcasted Hadamard product, and $\mathrm{SA}$, $\mathrm{CA}$, and $\mathrm{SC}$ are the outputs of the SAB, CAB, and SCB branches, respectively.

\paragraph{Spatial Attention Branch (SAB).} As shown in Figure~\ref{fig:SC2Head}\textcolor{cvprblue}{(a)}, for a given input feature, global average pooling and global max pooling are performed along the channel dimension to capture different semantic responses. The two responses are concatenated and aggregated to generate spatial attention weights, which are then multiplied with the input feature $\mathbf{X}$ to produce the output feature, expressed as:
\begin{equation}
\mathbf{SA}=f_{3 \times 3}^{s}\left(\left[f_{\text {spatial}}^{\text {Avg}}(\mathbf{X}), f_{\text {spatial}}^{\text {Max}}(\mathbf{X})\right] \odot \mathbf{X}\right),
\label{eq:SAB}
\end{equation}
where $f_{3 \times 3}^{s}$ represents a $3 \times 3$ convolution with a sigmoid activation function, and $f_{\text {spatial}}^{\text {Avg}}(\cdot)$ and $f_{\text {spatial}}^{\text {Max}}(\cdot)$ denote spatial average pooling and spatial max pooling, respectively.

\paragraph{Channel Attention Branch (CAB).} Each channel in an image feature map typically carries distinct semantic information, and assigning different weights to channels helps the network focus on the most informative ones. As shown in Figure~\ref{fig:SC2Head}\textcolor{cvprblue}{(b)}, for an input feature $\mathrm{X} \in \mathbb{R}^{C \times H \times W}$, channel-wise average pooling $f_{\text {channel}}^{\text {Avg}}(\cdot)$ and max pooling $f_{\text {channel}}^{\text {Max}}(\cdot)$ are applied using kernels of size $(H, W)$. The two pooled responses, corresponding to the same feature columns, are concatenated along the channel dimension to form a shared representation:
\begin{equation}
\mathrm{M}_{\mathrm{c}}=\left[f_{\text {channel}}^{\text {Avg}}(\mathbf{X}), f_{\text {channel}}^{\text {Max}}(\mathbf{X})\right].
\label{eq:CAB_Mc}
\end{equation}
The shared feature $\mathbf{M}_{\mathrm{c}} \in \mathbb{R}^{2C \times H \times W}$ is processed to enable feature interaction and split equally into two branches:
\begin{equation}
\mathbf{X}_{a}, \mathbf{X}_{m}=\operatorname{Chunk}_{2}\left(\operatorname{CBL}\left(\mathbf{M}_{c}\right)\right),
\label{eq:CAB_chunk}
\end{equation}
where $\operatorname{Chunk}_{2}(\cdot)$ divides the tensor into two equal parts along the channel dimension, and $\operatorname{CBL}(\cdot)$ denotes a subnetwork composed of a $1 \times 1$ convolution, batch normalization (BN), and LeakyReLU activation. The channel attention map $\mathrm{CA} \in \mathbb{R}^{C \times H \times W}$ is then computed as:
\begin{equation}
\mathrm{CA}=\mathbf{X} \odot \mathrm{F}_{\text {channel}}^{\text {Avg}}\left(\mathbf{X}_{a}\right) \odot \mathrm{F}_{\text {channel}}^{\text {Max}}\left(\mathbf{X}_{m}\right),
\label{eq:CAB_CA}
\end{equation}
where $\mathrm{F}_{\text {channel}}^{\text {Avg}}(\cdot)$ and $\mathrm{F}_{\text {channel}}^{\text {Max}}(\cdot)$ denote submodules within the channel attention mechanism, each consisting of convolution, BN, ReLU, another convolution, and Sigmoid.

\paragraph{Self-Calibration Branch.} As illustrated in Figure~\ref{fig:SC2Head}\textcolor{cvprblue}{(c)}, the SCB branch is designed to model contextual information effectively and establish long-range dependencies across spatial positions. For an input feature $\mathrm{X} \in \mathbb{R}^{C \times H \times W}$, the SCB computation is described as:
\begin{equation}
\mathbf{SC}=\delta_{s}\left(\mathbf{X}+\mathrm{B}_{2}\left(\operatorname{conv}\left(\mathrm{A}_{2}(\mathbf{X})\right)\right)\right),
\label{eq:SCB}
\end{equation}
where $\mathrm{A}_{2}(\cdot)$ denotes bilinear interpolation with an upsampling factor of 2, and $\mathrm{B}_{2}$ represents average pooling with a kernel size of $2 \times 2$ and stride 2. The resulting self-calibrated feature is then concatenated with the spatial and channel attention outputs to form the final fused feature representation for keypoint prediction.
\section{Experiments}

\begin{table*}[!t]
\centering
\caption{Quantitative results across different pose estimation baselines. HigherAssociativeEmbeddingHead is abbreviated as HAEHead. AssociativeEmbeddingHead is abbreviated as AEHead. }
\resizebox{\textwidth}{!}{
\begin{tabular}{lllcccccccc}
\toprule
\textbf{Methods} & \textbf{Backbone} & \textbf{Head} & \textbf{AP/\%} & \textbf{AP\textsubscript{50}/\%} & \textbf{AP\textsubscript{75}/\%} & \textbf{AR/\%} & \textbf{AR\textsubscript{50}/\%} & \textbf{AR\textsubscript{75}/\%} & \textbf{FLOPs/G} & \textbf{Params/M} \\

\midrule
DEKR~\cite{geng2021bottom}               & HRNet      & DEKRHead         & 87.2 & 95.8 & 90.3 & 89.0 & 96.7 & 91.9 & 44.416 & 29.548 \\
CID~\cite{wang2022contextual}               & HRNet      & CIDHead                     & 88.0 & 96.4 & 90.8 & 89.0 & 97.1 & 91.7 & 44.160 & 29.363 \\
CoupledEmbedding~\cite{wang2022regularizing}   & HRNet      & AEHead     & 72.2 & 90.5 & 75.4 & 78.0 & 95.0 & 77.2 & 41.100 & 28.641 \\
CoupledEmbedding~\cite{wang2022regularizing}   & HRNet      & HAEHead    & 73.9 & 90.1 & 74.0 & 80.4 & 96.6 & 82.5 & 40.500 & 28.541 \\
SimCC~\cite{li2022simcc}              & ResNet50   & SimCCHead                   & 87.4 & 96.0 & 91.0 & 89.9 & 96.7 & 92.9 & 5.493  & 36.753 \\
SimCC~\cite{li2022simcc}              & ResNet101  & SimCCHead                   & 87.4 & 97.0 & 91.6 & 89.8 & 97.5 & 91.7 & 9.140  & 55.745 \\
RTMPose~\cite{jiang2023rtmpose}            & CSPNext    & RTMCCHead                   & 88.6 & 97.0 & 90.6 & 89.0 & 97.5 & 92.7 & 1.926  & 13.550 \\
DWPose~\cite{yang2023effective}             & -          & -                           & 88.3 & 97.0 & 91.5 & 89.8 & 97.3 & 92.1 & 2.200  & -      \\
RTMO~\cite{lu2024rtmo}               & CSPDarknet & RTMOHead                    & 87.8 & 96.8 & 89.6 & 88.7 & 97.1 & 91.0 & 31.656 & 22.475 \\
\midrule
Ours (FSMC-Pose)               &  CowMountNet & SC2Head                    & \textbf{89.0} & \textbf{97.0} & \textbf{92.5} & \textbf{89.9} & \textbf{97.7} & \textbf{93.1} & \textbf{0.354} & \textbf{2.698} \\
\bottomrule
\end{tabular}
}
\label{tab:results_AP_AR}
\end{table*}

\subsection{Experimental Setup}
\paragraph{Evaluation Setting.}
The dataset in this study follows the COCO annotation format. To evaluate the similarity between predicted and ground-truth keypoints, we adopted the Object Keypoint Similarity (OKS) metric from the COCO dataset to calculate Average Precision (AP) and Average Recall (AR). The OKS is defined as follows:
\begin{equation*}
OKS = \frac{\sum_{i} \exp \left(-\frac{d_{i}^{2}}{2 s^{2} k_{i}^{2}}\right) \delta(v_{i}>0)}{\sum_{i} \delta(v_{i}>0)} \tag{13}
\end{equation*}
where $d_{i}$ denotes the Euclidean distance between the ground-truth and predicted keypoints, $v_{i}$ indicates the visibility flag of keypoint $i$, $s^{2}$ represents the object scale, and $k_{i}$ is the standard deviation corresponding to each keypoint, which varies depending on the annotation. $\delta(v_{i}>0)$ equals 1 when the keypoint is visible and 0 otherwise, meaning that only visible keypoints are considered in the computation; predictions for unannotated keypoints do not affect the final results. In addition, we also evaluate the model’s computational efficiency using parameters such as the total number of parameters and floating-point operations (GFLOPs).

\paragraph{Implementation Details.}
All models in this study were trained and tested on an Ubuntu 18.04 operating system using the PyTorch deep learning framework. The experiments were conducted on an NVIDIA Tesla P100 PCIe GPU with 16 GB of memory and an Intel Xeon E5-2680 v4 CPU running at 2.40 GHz. The PyTorch version used was 1.10.1 with CUDA 10.2 support. The Adam optimizer with a warm-up strategy was adopted. Considering both model complexity and dataset scale, the initial learning rate was set to 0.001, and the input image resolution was $256 \times 192$. Several data augmentation strategies were applied, including random scaling within a specified range, rotation, random horizontal shifting, and random occlusion of image regions. These augmentations introduce noise and variability into the data to prevent overfitting to specific image features or convergence to the local minima.

\begin{figure*}[t]
\centering
  \includegraphics[width=0.65\textwidth]{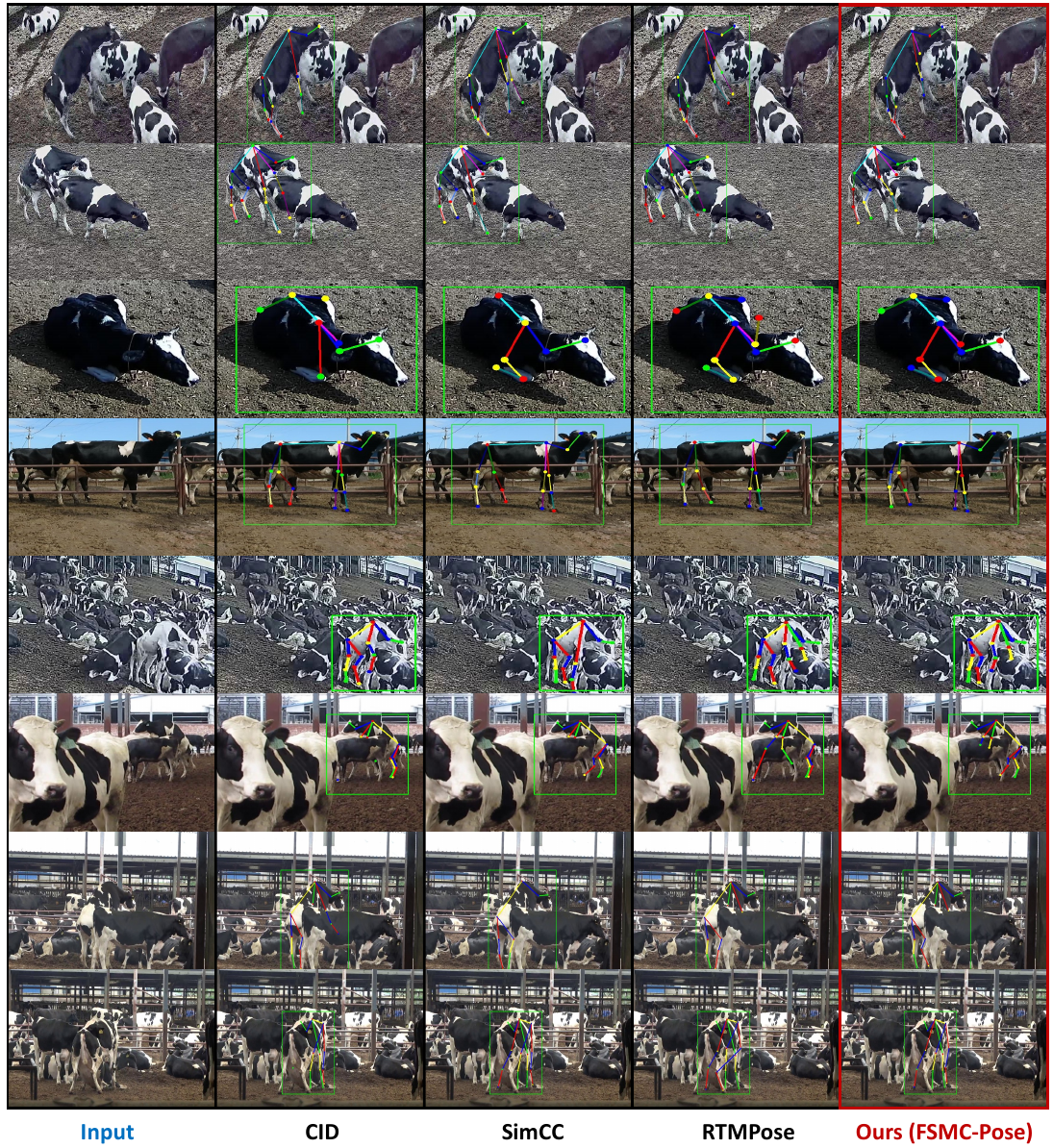}
\caption{Qualitative comparison of mounting pose estimation results on challenging real-world scenes. FSMC-Pose produces more accurate pose than CID~\cite{wang2022contextual}, SimCC~\cite{li2022simcc}, and RTMPose~\cite{jiang2023rtmpose}, especially under occlusion, cluttered backgrounds, and dense herd scenarios.}
\label{fig:results_visualization}
\end{figure*}
\begin{figure}[t]
\centering
  \includegraphics[width=\linewidth]{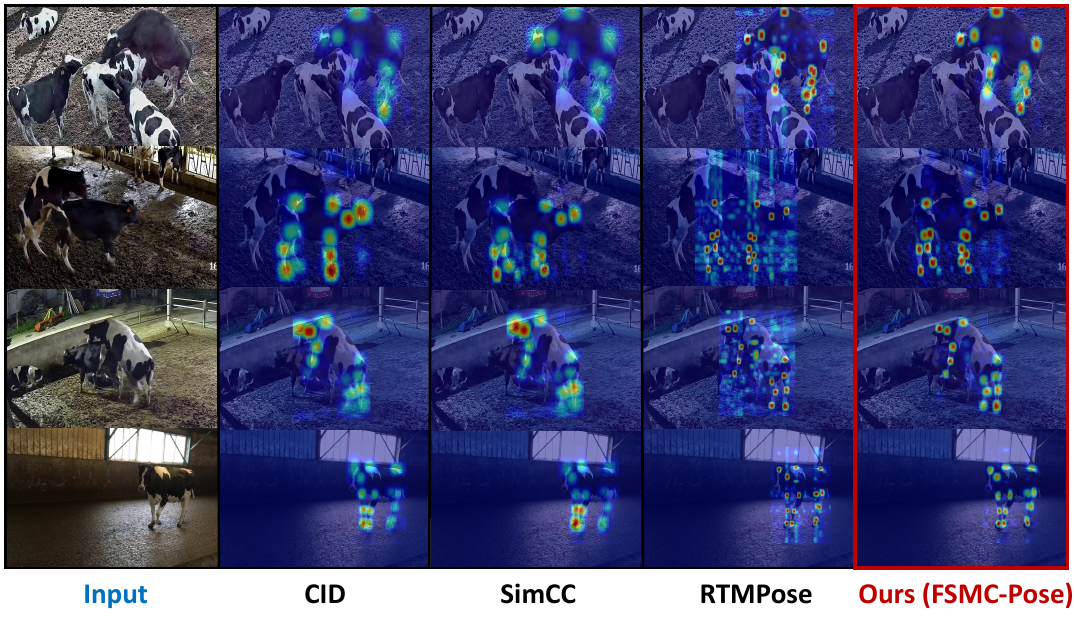}
\caption{Qualitative comparison of predicted keypoint heatmaps under complex herd scenes. FSMC-Pose produces more concentrated and well-localized responses, especially around limbs and joints, compared with other top-down methods~\cite{wang2022contextual,li2022simcc,jiang2023rtmpose}.}
\label{fig:Heatmaps}
\end{figure}

\subsection{Experimental Results}
\paragraph{Quantitative Results on Strong Baselines.}
We conducted a comprehensive evaluation of FSMC-Pose against several representative methods on the constructed dataset, including both top-down and bottom-up paradigms, to fully compare performance across different pose estimation approaches. The results are presented in Table~\ref{tab:results_AP_AR}, with the best results highlighted in bold. FSMC-Pose achieved the highest $89.0\%$ AP on this dataset, surpassing representative methods such as SimCC and RTMO, and outperforming other comparison models. This demonstrates the strong accuracy advantage of FSMC-Pose in the pose estimation task. Moreover, FSMC-Pose also achieved excellent results in AP\textsubscript{75}, AR\textsubscript{50}, and AR\textsubscript{75}, exceeding state-of-the-art methods such as SimCC and RTMPose, and significantly outperforming DEKR, CID, and RTMO. The results show that FSMC-Pose consistently extracts stable keypoint features across varying poses and scales, demonstrating strong generalization and robustness. Notably, FSMC-Pose also excels in model compactness and computational efficiency, with only 0.354M parameters and 2.698 GFLOPs. Compared with lightweight methods such as RTMPose and SimCC, it substantially reduces computational cost while maintaining high accuracy, achieving a balance between lightweight design and efficient inference.

\paragraph{Qualitative Results and Visualization.}
We qualitatively compare FSMC-Pose with representative top-down baselines (CID, SimCC, RTMPose) on both challenging test images and additional dense herd scenes, as illustrated in Figure~\ref{fig:results_visualization}. In the four groups of examples, cattle are densely packed, only a few keypoints of the mounting individual remain visible, and cloudy low-light conditions plus low camera resolution further increase difficulty. All compared models show obvious failures, including missing keypoints in heavily occluded regions and confused skeletons between overlapping individuals. In these settings, CID, SimCC, and RTMPose often exhibit missing or misplaced keypoints and confused skeleton assembly, especially when limbs of different cows overlap or when contrast is low. In contrast, FSMC-Pose consistently produces more coherent skeletons and more accurate keypoint localization. The frequency–spatial enhancement of SFEBlock and the multiscale modeling of RABlock help separate cattle from cluttered backgrounds and capture both small joints and large torso structures, while SC2Head stabilizes predictions under heavy overlap and partial occlusion. Overall, the visual results indicate that FSMC-Pose generalizes better to complex, real-world herd environments.

\paragraph{Qualitative analysis of keypoint heatmaps.}
We further compare heatmaps across models (Figure~\ref{fig:Heatmaps}). CID retains some localization ability but shows diffuse or blurred responses in cluttered areas, while SimCC struggles to separate individuals in crowded scenes, leading to displaced keypoints. FSMC-Pose generates more compact and well-aligned heatmaps, especially around limbs and joints.

\subsection{Ablation Study}
\paragraph{Effect of each module.}
We evaluate the effectiveness of the proposed modules using RTMPose~\cite{jiang2023rtmpose} as the baseline, results are summarized in Table~\ref{tab:ablation}. The baseline has $13.55$M parameters and strong accuracy but high deployment cost. Replacing the backbone with a lightweight MobileNet~\cite{sandler2018mobilenetv2} reduces parameters to $1.609$M (Table~\ref{tab:ablation}) at the expense of accuracy, suggesting the need for auxiliary modules to recover representation capacity. On top of this lightweight design, we introduce three modules (SFEBlock, RABlock, and SC2Head) for trade-off accuracy and efficiency. Individually, SFEBlock enhances edge and texture cues; RABlock enlarges the receptive field to strengthen global structure modeling; SC2Head applies spatial–channel attention that improves recall but may slightly reduce localization precision when used alone. Combinations further boost performance: SFEBlock+RABlock balances fine detail and global context, RABlock+SC2Head couples receptive-field expansion with attention to emphasize salient keypoints, and SFEBlock+SC2Head shows complementary gains with minimal overhead. The full model with all three modules attains the best overall accuracy and efficiency trade-off while keeping the parameter count compact.

\begin{table}[tpb]
\centering
\caption{Ablation study of the proposed modules. We ablate SFEBlock, RABlock, and SC2Head individually and in combination. The baseline is RTMPose~\cite{jiang2023rtmpose} with CSPNext backbone. The best results are highlighted in \textbf{bold}.}
\resizebox{\linewidth}{!}{
\begin{tabular}{cccccccc}
\toprule
\multicolumn{2}{c}{\textbf{CattleMountNet}} & \multicolumn{1}{c}{\textbf{Head}} & \multicolumn{4}{c}{\textbf{Metrics (\%)}} & \multirow{2}{*}{\textbf{Params/M}} \\ 
\cmidrule(lr){1-2} \cmidrule(lr){3-3} \cmidrule(lr){4-7}
\textbf{SFEBlock} & \textbf{RABlock} & \textbf{SC2Head} & \textbf{AP} & \textbf{AP\textsubscript{75}} & \textbf{AR} & \textbf{AR\textsubscript{75}} & \\
\midrule
\multicolumn{3}{l}{Baseline} & 88.4 & 90.6 & 89.0 & 92.7 & 13.55 \\
\multicolumn{3}{l}{Baseline w/ MobileNet} & 87.8 & 89.5 & 89.0 & 91.3 & 1.609 \\
\checkmark &  &  & 88.2 & 91.7 & 89.6 & 92.3 & 1.903 \\
 & \checkmark &  & 88.0 & 90.8 & 89.0 & 91.7 & 2.393 \\
 &  & \checkmark & 87.5 & 90.7 & 89.3 & 91.9 & 1.620 \\
\checkmark & \checkmark &  & 88.7 & 91.6 & 89.7 & 91.5 & 2.687 \\
 & \checkmark & \checkmark & 88.3 & 91.8 & 89.9 & 91.9 & 2.404 \\
\checkmark &  & \checkmark & 88.6 & 92.1 & 89.8 & 92.1 & 1.914 \\
\checkmark & \checkmark & \checkmark & \textbf{89.0} & \textbf{92.5} & \textbf{89.9} & \textbf{93.1} & 2.698 \\
\bottomrule
\end{tabular}
}
\label{tab:ablation}
\end{table}

\begin{table}[t]
\centering
\caption{Comparison of different attention mechanisms. Channel-only: CSA and ECA. Spatial-only: SAM and EMA. Joint spatial–channel: CBAM, SCAM and GCSA. Best performance is highlighted in \textbf{bold}.}
\resizebox{\linewidth}{!}{
\begin{tabular}{ccccc}
\toprule
\textbf{Attention Mechanisms} & \textbf{AP/\%} & \textbf{AP\textsubscript{75}/\%} & \textbf{AR/\%} & \textbf{AR\textsubscript{75}/\%} \\

\midrule
CSA   & 88.5 & 91.4 & 89.4 & 92.1 \\
ECA   & 88.1 & 90.3 & 89.3 & 91.0 \\
\midrule
SAM   & 88.4 & 91.6 & 89.1 & 91.3 \\
EMA   & 88.1 & 89.0 & 89.2 & 90.5 \\
\midrule
CBAM  & 88.4 & 90.8 & 89.4 & 91.2 \\
SCAM  & 88.2 & 90.4 & 89.6 & 91.7 \\
GCSA  & 88.3 & 90.5 & 89.5 & 91.3 \\
\midrule
SC2Head  & \textbf{89.0} & \textbf{92.5} & \textbf{89.9} & \textbf{93.1} \\
\bottomrule
\end{tabular}
}
\label{tab:attention}
\end{table}

\paragraph{Effect of SC2Head Attention.}
We evaluate SC2Head against several representative attention mechanisms, including channel-only (CSA, ECA), spatial-only (SAM, EMA), and joint spatial–channel attention (CBAM, SCAM, GCSA). To ensure a fair comparison, all attention modules are embedded in the same position as SC2Head, with identical backbone structures, training strategies, input resolutions, and datasets, and only the attention module type is replaced. The experimental results are shown in Table~\ref{tab:attention}. The results show that SC2Head achieves an AP of $89.0\%$ and an AR of $89.9\%$, improving AP by $0.5$ over CSA and $0.6$ over CBAM, and also yields the highest AR\textsubscript{75} ($93.1\%$), indicating its excellent performance under occlusion and low-contrast conditions. These gains stem from its spatial–channel co-calibration with a self-calibration branch that dynamically couples channel semantics and spatial responses, forming an efficient pathway for enhancing discriminative keypoint features.

\subsection{Comparison of Inference Speed}
To further evaluate the deployment potential of FSMC-Pose, we compare inference speed with several mainstream pose estimation methods under identical hardware and experimental conditions, as summarized in Table~\ref{tab:speed}. FSMC-Pose achieves $216.58$ FPS and requires only $0.354$ GFLOPs and $2.698$M parameters, clearly outperforming DEKR~\cite{geng2021bottom}, CoupledEmbedding~\cite{wang2022regularizing}, RTMO~\cite{lu2024rtmo}, and the real-time oriented CID~\cite{wang2022contextual} in terms of both speed and model complexity. The slightly smaller than expected speed gap over CID is mainly due to the use of several complex modules with non-standard convolutions and frequent tensor reshaping, which introduce additional memory and scheduling overhead. Even so, FSMC-Pose still provides real-time inference at over $200$ FPS with significantly reduced computation and parameters, making it well suited for edge devices and latency-sensitive applications where efficiency, resource consumption, and deployment cost are critical.

\begin{table}[tpb]
\centering
\caption{Comparison of inference speed. FLOPs/G, Params/M, and FPS denote computation, model size, and runtime speed, respectively. Best results are highlighted in \textbf{bold}.}
\resizebox{=0.85\linewidth}{!}{
\begin{tabular}{cccc}
\toprule
\textbf{Methods} & \textbf{FLOPs/G} & \textbf{Params/M} & \textbf{FPS} \\
\midrule
DEKR               & 8.328 & 29.548 & 37.57  \\
CID                & 8.093 & 29.363 & 184.09 \\
CoupledEmbedding   & 7.590 & 28.541 & 89.90  \\
RTMO               & 31.656 & 22.475 & 78.23  \\
\midrule
Ours (FSMC-Pose)               & \textbf{0.354} & \textbf{2.698} & \textbf{216.58} \\
\bottomrule
\end{tabular}
}
\label{tab:speed}
\end{table}

\section{Conclusion and Discussion}
In this work we address mounting pose estimation for dairy cattle in cluttered herd environments, a setting largely overlooked in animal pose estimation. We propose FSMC-Pose, a lightweight top-down framework that couples the CattleMountNet backbone with SFEBlock and RABlock for frequency–spatial fusion and multiscale aggregation, and SC2Head for spatial–channel self-calibration under inter-animal overlap. On a benchmark combining the self-collected MOUNT-Cattle and public NWAFU-Cattle dataset, FSMC-Pose outperforms baselines in AP and AR while retaining low computational cost and real-time inference on commodity GPUs. Extensive Qualitative visualizations further show that FSMC-Pose generalizes well to complex farm scenes and yields robust mounting pose estimation. In future work, we plan to extend mounting analysis to full estrus behavior pipelines by integrating temporal cues and multi-camera views and to explore large-scale deployment in precision livestock farming.


\section*{Acknowledgement}
This work was supported by the Beijing Natural Science Foundation (No.~4242037), the Youth Project of MOE (Ministry of Education) Foundation on Humanities and Social Sciences (No.~23YJCZH223), and the National Natural Science Foundation of China (No.~72501020,~U2568225).



    \small
    \bibliographystyle{ieeenat_fullname}
    \bibliography{main}


\end{document}